\documentclass[10pt,a4paper,twoside,twocolumn]{article}

\usepackage[final]{cmpproc}       

\def\cvwwPaperID{33}

\usepackage{graphicx}
\usepackage{hyperref}

\usepackage{amsmath,amssymb,amsfonts,amsthm,color,subfigure}
\usepackage{ifpdf}
\usepackage{multirow}
\usepackage{booktabs}
\usepackage{nicefrac}
\usepackage{xcolor}
\usepackage{xspace}
\usepackage{enumerate}

\usepackage[section]{algorithm}
\usepackage{algorithmic}

\theoremstyle{remark}
\newtheorem{remark}{Remark}

\newcommand{\scal}[2]{\left\langle #1,#2 \right\rangle}
\def\Carre#1#2{\vbox{
   \hrule height .#2pt
   \hbox{\vrule width .#2pt height #1pt \kern #1pt
      \vrule width .#2pt}
   \hrule height .#2pt}}

\def\R{\mathbb{R}}

\def\arg{\textup{arg}\,}

\newcommand{\norm}[2][]{\|{#2}\|_{#1}}

\newcommand{\eps}{\varepsilon}

\newcommand{\cC}{\mathcal{C}}

\newcommand{\cI}{\mathcal{I}}

\newcommand{\cR}{\mathcal{R}}

\newcommand{\diag}{\mathrm{diag}}

\begin{document}

\title{A bi-level view of inpainting - based image compression}
\subtitle{}
\author{Chen, Yunjin; Ranftl, Ren{\'e}; Pock, Thomas}
\affiliation{Institute for Computer Graphics and Vision,\\Graz University of Technology, Austria\\
   \texttt{\href{mailto:cheny@icg.tugraz.at}{\{cheny,ranftl,pock\}@icg.tugraz.at}}}

\begin{proceeding}

\begin{abstract}
Inpainting based image compression approaches, especially linear and non-linear diffusion models, are an 
active research topic for lossy image compression. The major challenge
in these compression models is to find a small set of descriptive
supporting points, which allow for an accurate reconstruction of the
original image. 
It turns out in practice that this is a challenging problem even for the simplest Laplacian interpolation  
model. In this paper, we revisit the Laplacian interpolation compression model and introduce two fast algorithms, namely 
successive preconditioning primal dual algorithm and the recently proposed iPiano algorithm, to solve this problem efficiently. 
Furthermore, we extend the Laplacian interpolation based
compression model to a more general form, which is based on principles
from bi-level optimization. We investigate two different variants of the Laplacian model, namely biharmonic interpolation and 
smoothed Total Variation regularization. Our numerical results show that 
significant improvements can be obtained from the biharmonic interpolation model, and it can recover an image with very high 
quality from only 5\% pixels. 
\end{abstract}

\section{Introduction}
Image compression is the task of storing image data in a compact form by reducing irrelevance and redundancy of 
the original image. Image compression methods roughly fall into two main types: lossless compression and lossy compression. In 
this paper, we focus on lossy compression methods. The objective of
lossy compression methods is to reduce the original image data as much
as possible while still providing a visually acceptable reconstruction
from the compressed data. 
Lossy image compression can be handled with two different approaches:
(1) reducing the data in the original image domain, i.e. by removing a
majority of the image pixels; (2) reducing data in a transform
domain, such as Discrete cosine transform (DCT) or Wavelet transform.
The remaining data (compressed data) is used to reconstruct the original 
image. It is well known that the former approach is named as image inpainting in the literature \cite{criminisi2004region, 
bertalmio2000image, shen2003inpainting}, and 
the latter strategy is exploited in the currently widely used standard image compression techniques such as JPEG and JPEG2000 
\cite{pennebaker1992jpeg, taubman2002jpeg2000}. 
In this paper, we focus on the strategy of reducing the data in the image 
domain and then recovering an image from a few data points, i.e., image inpainting. 

There are thousands of publications studying the topic of image inpainting in the literature, see e.g., \cite{
criminisi2004region, bertalmio2000image, shen2003inpainting, dobrosotskaya2008wavelet} and references therein. 
In most cases, one does not have influence on the chosen data
points. In the context of image inpainting, one usually randomly selects a specific 
amount of pixels which act as supporting points for the inpainting
model, e.g., 5\%. 
In order to get high quality reconstructions in such a scenario, one
has to rely on sophisticated inpainting models. 
However, the task of image inpainting is to recover an image from only 
a few observations, and therefore, if the randomly selected data points do not carry sufficient information of the original image, 
even sophisticated inpainting models will fail to provide an accurate
reconstruction.

This observation motivated researchers to consider a different
strategy for building inpainting based compression models, i.e. to
find the optimal data points required for inpainting, given a specific
inpainting model. Prior work in this direction can be found in 
\cite{GalicWWBBS08, MainbergerSpatialTonal, belhachmi2009choose, SchmaltzWB09, HoeltgenSW13, MainbergerBWF11}.
Belhachmi \textit{et al.} \cite{belhachmi2009choose} propose an analytic approach 
to choose optimal interpolation data for Laplacian interpolation, based on the modulus of the Laplacian. The work in \cite{MainbergerSpatialTonal} 
demonstrates that carefully selected data points can result in a significant improvement of the reconstruction quality 
based on the same Laplacian interpolation, when compared to the prior work \cite{belhachmi2009choose}. However, 
this approach takes millions of iterations to converge and therefore
is very time consuming. The very recent work \cite{HoeltgenSW13} 
pushed forward this research topic, where the task of finding optimal data for Laplacian interpolation was 
explicitly formulated as an optimization problem, which was solved by
a successive primal dual algorithm. While their work 
still requires thousands of iterations to reach a meaningful solution, this new 
model {shed light on the possibility of employing optimization
approaches} and shows state-of-the-art 
performance for the problem of finding optimal data points for
inpainting based image compression. 

The work of \cite{HoeltgenSW13} is the starting point of this paper. In this paper, we extend the model of finding optimal 
data for Laplacian interpolation to a more general model, which comprises the model in \cite{HoeltgenSW13} as a special case. 
We introduce two novel models to improve the compression performance,
i.e., to get better reconstruction quality with the same amount of
pixels. Finally, we introduce efficient algorithms to solve the corresponding optimization problems. Namely, we make 
the following two main contributions in this paper: 

(1) We comprehensively 
investigate two efficient algorithms, which can be applied to solve the corresponding optimization problems, including 
successive preconditioning primal dual \cite{Pock2011} and a recently
published algorithm for non-convex optimization - iPiano \cite{ipiano}. 

(2) We explore two variants of Laplacian interpolation based image compression 
to improve the compression performance, namely, a model
based on the 
smoothed TV regularized inpainting model and biharmonic interpolation. It turns out that biharmonic 
interpolation can lead to significant improvements
over Laplacian interpolation.

\section{Extension of the Laplacian interpolation based image compression model} 
The original Laplacian interpolation is formulated as the following boundary value problem: 
\begin{align}\label{constlap}
- \Delta u = 0, &\quad\text{on} ~\Omega ~\texttt{\char`\\} ~ I\nonumber\\
u = g, &\quad\text{on} ~ I\\
\partial_n u = 0, &\quad\text{on} ~\partial \Omega ~\texttt{\char`\\} \partial I\,, \nonumber
\end{align}
where $g$ is a smooth function on a bounded domain $\Omega \subset \R^n$ with regular boundary $\partial \Omega$. 
The subset $I \subset \Omega$ denotes the set with known observations and $\partial_n u$ denotes the gradient of 
$u$ at the boundary. $\Delta$ denotes the Laplacian operator. 

It is shown in \cite{MainbergerSpatialTonal, HoeltgenSW13} that the problem \eqref{constlap} is equivalent to 
the following equation
\begin{align}\label{constlap2}
c(x)(u(x) - g(x)) - (1-c(x))\Delta u(x) = 0, &\quad\text{on} ~\Omega \\
\partial_n u(x) = 0, &\quad\text{on} ~\partial \Omega ~\texttt{\char`\\} ~\partial I\,,\nonumber
\end{align}
where $c$ is the indicator function of the set $I$, i.e., $c(x) = 1$, if $x \in I$ and $c(x) = 0$ elsewhere. 
By using the Neumann boundary condition, the discrete form of \eqref{constlap2} is given by 
\begin{equation}\label{laplacian}
C(u - g) - (\cI - C)\Delta u = 0\,,
\end{equation}
where the input image $g$ and the reconstructed image $u$ are vectorized to column vectors, i.e., $g \in \R^N$ and 
$u \in \R^N$, $C = \diag(c) \in \R^{N \times N}$ is a diagonal matrix with the vector $c$ on its main diagonal, 
$\Delta \in \R^{N \times N}$ is the Laplacian operator and $\cI$ is the identity matrix. 
The underlying philosophy behind this model is to inpaint the region ($\Omega ~\texttt{\char`\\}~ I$) 
by using the given data in region $I$, such that the recovered image is second-order smooth in the inpainting region, 
i.e., $\Delta u = 0$. 

Note that the inpainting mask $c$ in \eqref{laplacian} is binary. However, 
as shown in \cite{HoeltgenSW13}, equation \eqref{laplacian} still makes sense when 
$c$ is relaxed to a continuous domain such as $\R$. Due to this observation, the task of finding optimal 
interpolation data can be explicitly formulated as the following optimization problem:
\begin{align}\label{mask}
  \min_{u,c} & \frac{1}{2}\|u - g\|_2^2 + \lambda\|c\|_1\\
  \text{s.t.}  \;  & C(u - g) - (\cI -C)\Delta u = 0\,, \nonumber
\end{align}
where the parameter $\lambda$ is used to control the percentage of pixels used for inpainting. 
When $\lambda = 0$, the optimal solution of \eqref{mask} is $c \equiv 1$, i.e., all the pixels are used; 
when $\lambda = \infty$, the optimal solution is $c \equiv 0$, i.e.,
none of the pixel are used. 
Compared to the original formulation in \cite{HoeltgenSW13}, we omit a very small quadratic term 
$\frac \eps 2 \|c\|^2_2$, because we found that it is not necessary in practice. 

Observe that if $c \in \cC = [0, 1)^N$, we can multiply the constraint equation in \eqref{mask} by a diagonal positive-definite matrix 
$(\cI - C)^{-1}$, which results in
\begin{equation}\label{constraint}
B(c) (u - g) - \Delta u = 0 \,,
\end{equation}
where $B(c) = \diag(c_1/(1-c_1), \cdots, c_N/(1-c_N))$. It is clear that the constraint equation \eqref{constraint} can 
be equivalently formulated as the following minimization problem
\begin{equation}\label{loweropt}
u(c) = \arg \min_{u} \frac 1 2 \|\nabla u\|^2_2 + \frac 1 2 \|B(c)^{\frac 1 2} (u - g)\|^2_2\,,
\end{equation}
where $\nabla$ is the gradient operator, and  $\Delta = - \nabla^\top\nabla$. 
Therefore, it turns out that the Laplacian interpolation is exactly the Tikhonov regularization technique for image inpainting, 
where the first term can be seen as the regularization term based on the gradient operator, and the second term as 
the data fidelity term. 

Now, let us consider how to improve the performance of the regularization based inpainting model \eqref{loweropt}. 
The only thing we can change is the regularization term. There are 
two possible directions: (1) considering higher-order linear operators, e.g., $\Delta$, to replace the 
first-order derivative operator $\nabla$; 
(2) replacing the quadratic regularization with more 
robust penalty functions, such as $\ell_p$ quasi-norm with $p \in (0, 1]$. 

The linear operators $\nabla$ and $\Delta$ can be interpreted as linear filters, the corresponding 
linear filters are shown in Figure~\ref{fig:derivatives}. If we make use of $\nabla$ in the inpainting model \eqref{loweropt}, 
the resulting operator $\Delta$ makes the inpainting process only involve the information from its nearest neighborhood; 
however, if we turn to the $\Delta$ operator, the resulting operator $\Delta^2$ (biharmonic operator) 
can involve more information from larger neighborhood, see Figure~\ref{fig:derivatives}. 
In principle, this should bring some improvement of inpainting 
performance; besides this, the biharmonic operator is mathematically
meaningful in itself, implying higher-order smoothness of the solution $u$.

\begin{figure}[t!]
\begin{center}
  {\includegraphics[width=0.15\linewidth]{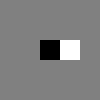}}
  {\includegraphics[width=0.15\linewidth]{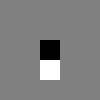}}
  {\includegraphics[width=0.15\linewidth]{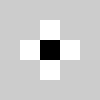}}
  {\includegraphics[width=0.15\linewidth]{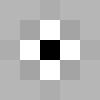}}
\end{center}
\caption{Linear operators shown as filters of size $5 \times 5$: 
from \textbf{left} to \textbf{right}, 
$\nabla_x$, $\nabla_y$, $\Delta$ and biharmonic operator ($\Delta^2$)
}
\label{fig:derivatives}
\end{figure}

Regarding the penalty function, quadratic function is known to generate over smooth results, especially for edges, 
and therefore many other edge-aware penalty functions  have been proposed. 
A straightforward extension is to make use of the $\ell_1$ norm, which leads to the well-known Total Variation (TV) 
regularization (still convex model). Since exact TV regularization suffers from the 
drawback of piece-wise constant solutions, we employ the following
smoothed version of TV regularization, which is parameterized  by a
small smoothing parameter $\eps$:
\[
\|\nabla u\|_\eps = \sum_{i=1}^N \sqrt{(\nabla_x u)_i^2 + (\nabla_y u)_i^2 + \eps^2} \,,
\]
where $\nabla_x u$ and $\nabla_y u$ denote the gradient in $x$ direction and $y$ direction, respectively. 
We will show in the next section that this smooth technique is also necessary for optimization. 

Using these considerations, we arrive at a general formulation of the inpainting-based image compression model, which is given by 
the following bi-level optimization problem:
\begin{align}\label{general}
  \min_{c \in \cC} & \frac{1}{2}\|u(c) - g\|_2^2 + \lambda\|c\|_1\\
  \text{s.t.}  \;  & u(c) = \arg \min_{u} \cR(u) + \frac 1 2 \|B(c)^{\frac 1 2} (u - g)\|^2_2\,, 
\nonumber
\end{align}
where the upper level problem is defined as the trade-off between the sparsity of the chosen data and the 
reconstruction quality, while the lower-level problem is given as the regularization based inpainting model. In the 
lower-level problem, $\cR(u)$ defines a regularization on $u$, and in this paper we investigate three different regularizers
\begin{equation}
\cR(u) = 
\begin{cases}
\frac 1 2 \|\nabla u\|_2^2 & \text{Laplacian interpolation} \\
\frac 1 2 \|\Delta u\|_2^2 & \text{Biharmonic interpolation}\\
\|\nabla u\|_\eps & \text{Smoothed TV regularization}
\end{cases}
\end{equation}

\section{Efficient algorithms for solving inpainting based image compression problems}
In the prior work \cite{HoeltgenSW13}, a successive primal dual
algorithm was used in order to solve 
the Laplacian interpolation based image compression problem \eqref{general}, where tens of thousands inner iterations and 
thousands of outer iterations are required to reach convergence. 
Since this is too time consuming for practical applications, we first
investigate efficient algorithms to solve
problem \eqref{general}. 
\subsection{Successive Preconditioning Primal Dual algorithm (SPPD)}
A straightforward method to accelerate the algorithm in \cite{HoeltgenSW13} is to make use of the diagonal 
preconditioning technique \cite{Pock2011} for the inner primal dual algorithm, while keeping the outer iterate unchanged. 
The basic principle of the successive primal dual algorithm,
is to linearize the constraint of~\eqref{general}, i.e., the lower-level problem. 
For smooth regularization terms $\cR(u)$, the lower-level problem of
\eqref{general} can be equivalently written using its first-order
optimality conditions:
\begin{equation}\label{lowereq}
T(u,c) = \frac {\partial \cR(u)} {\partial u} + B(c) (u - g) = 0.
\end{equation}
Using Taylor expansion, we linearize \eqref{lowereq} around a point~$(\hat u, \hat c)$:
\begin{equation}\label{linearize}
T(u,c) \approx T(\hat u, \hat c) + \left(\frac {\partial T}{\partial u}\big|_{\hat u}\right)^\top (u - \hat u) + 
\left(\frac {\partial T}{\partial c}\big|_{\hat c}\right)^\top (c - \hat c) = 0. 
\end{equation}
Let $(\hat u, \hat c)$ be a feasible point of constraint \eqref{lowereq}, i.e., $T(\hat u, \hat c) = 0$, 
and substitute the linearized 
constraint back into the initial problem \eqref{general}, we arrive at the following constrained optimization problem
\begin{align}\label{pd}
  \min_{c \in \cC, u} & \frac{1}{2}\|u - g\|_2^2 + \lambda\|c\|_1 + \frac {\mu_1} 2 \|c - \hat c\|^2_2 + 
\frac {\mu_2} 2 \|u - \hat u\|^2_2 \nonumber \\
  \text{s.t.}  \;  & D_u u + D_c c + q = 0\,,
\end{align}
where $D_u = \left(\frac {\partial T}{\partial u}\big|_{\hat u}\right)^\top$, 
$D_c = \left(\frac {\partial T}{\partial c}\big|_{\hat c}\right)^\top$, 
$q = - D_u \hat u - D_c \hat c$. 
Note that the linearized constraint is only valid around a small neighborhood of $(\hat u, \hat c)$, and therefore 
we have to add two additional penalty term $\frac {\mu_1} 2 \|c - \hat c\|^2_2$ and $\frac {\mu_2} 2 \|u - \hat u\|^2_2$ 
to ensure that the solution $(u^*,c^*)$ is in the vicinity of $(\hat u, \hat c)$. The saddle-point formulation of 
\eqref{pd} is written as
\begin{align}\label{saddlepoint}
\max_{p} \min_{(u,c)} \scal{K \binom{u}{c} + q}{p} + 
\frac{1}{2}\|u - g\|_2^2 + \lambda\|c\|_1 + \nonumber \\
\frac {\mu_1} 2 \|c - \hat c\|^2_2 + 
\frac {\mu_2} 2 \|u - \hat u\|^2_2 + \delta_\cC(c) \,,
\end{align}
where $K = (D_u, D_c)$, $\delta_\cC(c)$ is the indicator function of set~$\cC$, and $p \in \R^N$ is the Lagrange multiplier 
associated with the equality constraint in \eqref{pd}.

\begin{remark}\label{remark1}
Note that for Laplacian and biharmonic interpolation, we do not
restrict $c$ to the set $\cC$, and we make use of 
the original constraint in \eqref{mask}, i.e., 
\[
C(u - g) - (\cI - C)L u = 0 \,, 
\]
where $L = -\Delta$ for Laplacian interpolation, and ${L = - \Delta^2}$ for biharmonic interpolation. 
Therefore, the indicator function $\delta_\cC(c)$ in equation
\eqref{saddlepoint} can be dropped for these models. However, for the TV regularized model or other 
possible regularization techniques, we have to strictly rely on \eqref{saddlepoint}.
\end{remark}
\begin{remark}\label{remark2}
It was stated in previous work \cite{HoeltgenSW13} that there is no need to introduce an additional penalty term for 
variable $u$, because $u$ continuously depends on $c$. However, we find that for biharmonic interpolation, 
we have to keep the penalty term for $u$, otherwise, the resulting algorithm will suffer from zigzag behavior 
when it gets close to the optimal solution. 
\end{remark}
It is easy to work out the Jacobi matrices $D_u$ and $D_c$ for Laplacian and biharmonic interpolation, which are given as
\[
\begin{cases}
D_u(\hat u, \hat c) = \diag(\hat c) - (\cI - \diag(\hat c)) L, \\
D_c(\hat u, \hat c) = \diag(\hat u - g + L \hat u ) \,. 
\end{cases}
\]
For smooth TV regularized inpainting model, the constraint~\eqref{lowereq} is written as 
\[
\nabla^\top\cdot \binom{\frac {\nabla_x u} {\rho}}{\frac {\nabla_x u} {\rho}} + B (u - g) = 0 \,,
\]
where $\rho = \sqrt{\nabla_x^2 u + \nabla_y^2 u + \eps^2}$, and $\nabla = \binom{\nabla_x}{\nabla_y}$. 
The Jacobi matrices $D_u$ and $D_c$ are given by
\begin{align}\label{hessian}
\begin{cases}
D_c(\hat u, \hat c) = \diag(\frac {1} {(1-c)^2}) \cdot \diag(u - g), \\ 
D_u(\hat u, \hat c) = 
\binom{\nabla_x}{\nabla_y}^\top \cdot
\diag \binom{\frac {\nabla_y^2 u + \eps^2}{\rho^3}}{\frac {\nabla_x^2 u + \eps^2}{\rho^3}}
\cdot \binom{\nabla_x}{\nabla_y} - \\
\hspace{1.75cm} \binom{\nabla_y}{\nabla_x}^\top \cdot
\diag \binom{\frac {\nabla_x u \odot \nabla_y u}{\rho^3}}{\frac {\nabla_x u \odot \nabla_y u}{\rho^3}}
\cdot \binom{\nabla_x}{\nabla_y} + B\,,
\end{cases}
\end{align}
where $\odot$ denotes point-wise multiplication. 

We make use of the diagonal preconditioning technique of \cite{Pock2011} to choose the 
preconditioning matrices $\Gamma$ and $\Sigma$. 
\[
\Gamma = \diag(\tau), ~\Sigma = \diag(\sigma) \,,
\]
where $\tau_j = \frac 1 {\sum_{i = 1}^{N}|K_{i,j}|^{2-\gamma}}, 
\sigma_i = \frac 1 {\sum_{j = 1}^{2N}|K_{i,j}|^{\gamma}}$. 
The we employ the preconditioning primal dual Algorithm~\ref{algo1} to solve problem \eqref{saddlepoint}.

\begin{algorithm}\caption{Preconditioning PD for solving problem \eqref{saddlepoint}}\label{algo1}
\begin{itemize}
\item[(1)] 
Compute the preconditioning matrices $\Gamma$ and $\Sigma$ and choose 
$\theta \in [0, 1]$
\item[(2)] Initialize $(u,c)$ with $(\hat u, \hat c)$, and $\bar{p} = 0$. 
\item[(3)] 
Then for $k \geq 0$, update $(u^k,c^k)$ and $p^k$ as follows:
\begin{equation}\label{update}
\begin{cases}
p^{k+1} = p^k + \Sigma \left( K\binom{u^k}{c^k} + q \right) \\
\bar{p}^{k+1} = p^{k+1} + \theta(p^{k+1} - p^{k}) \\
\binom{u^{k+1}}{c^{k+1}} = 
(\cI + \Gamma \partial G)^{-1} \left(\binom{u^k}{c^k} - \Gamma K^\top\bar{p}^{k+1}\right) 
\end{cases}
\end{equation}
\end{itemize}
\end{algorithm}
For Laplacian and biharmonic interpolation, the function $G(u,c)$ in \eqref{update} is given as
\[
G(u,c) = \frac{1}{2}\|u - g\|_2^2 + \frac {\mu_2} 2 \|u - \hat u\|^2_2
+ \lambda\|c\|_1 + \frac {\mu_1} 2 \|c - \hat c\|^2_2. 
\]
It turns out that the proximal map with respect to $G$ simply poses point-wise operations, which is given as
\begin{align}\label{guc}
\binom{u}{c} = (\cI + \Gamma \partial G)^{-1}\binom{\tilde{u}}{\tilde{c}} \Longleftrightarrow \nonumber \\
\begin{cases}
u_i =  \frac {\tilde{u}_i + \tau_i^1 g_i + \mu_2\tau_i^1 \hat u_i}{1 + \tau_i^1 + \mu_2 \tau_i^1} 
\quad i = 1 \cdots N\\
c_i = \text{shrink}_{\frac{\lambda \tau_i^2}{1 + \tau_i^2 \mu_1}}\left(
\frac{\tilde{c}_i + \tau_i^2 \mu_1 \hat c_i}{1 + \tau_i^2 \mu_1}
\right) \,,
\end{cases}
\end{align}
where the soft shrinkage operator is given by $\text{shrink}_\alpha(x) = \text{sgn}(x) \cdot \max (|x| - \alpha, 0)$, 
and $\tau = \binom{\tau^1}{\tau^2}$. 

For smooth TV regularization, the function $G$ is given by
\[
G(u,c) = \frac{1}{2}\|u - g\|_2^2 + \frac {\mu_2} 2 \|u - \hat u\|^2_2
+ \lambda \sum_{i=1}^{N} c_i + \frac {\mu_1} 2 \|c - \hat c\|^2_2 + \delta_\cC(c). 
\]
The proximal map for $u$ is the same as in \eqref{guc}, the solution
for $c$ can be computed by 
\[
c_i = \text{Proj}_\cC \left(\frac{\tilde{c}_i + \tau_i^2 \mu_1 \hat c_i - \tau_i^2 \lambda}{1 + \tau_i^2 \mu_1}
\right)
\]
\subsection{iPiano}
Observe that in problem \eqref{general} the lower-level problem can be
solved for $u$, and the result can be substituted back 
into the upper-level problem. It turns out that this results in an
optimization problem which only depends on the variable $c$. 
It is demonstrated in our previous work \cite{ipiano} that this optimization problem can be solved efficiently by using 
the recently proposed algorithm - iPiano. Our experiments will show
that this strategy is more efficient than the successive
preconditioning primal dual algorithm.

For Laplacian and biharmonic interpolation, we can solve $u$ in closed form, 
i.e., $u = A^{-1}Cg$. This results in the following optimization 
problem, which only depends on variable $c$: 
\begin{equation}\label{newmask}
  \min_{c}\frac 12\| A^{-1}\diag(c) g - g\|_2^2 + \lambda\|c\|_1\,,
\end{equation}
where $A = C + (C - \cI) L$. Casting \eqref{newmask} in the form of iPiano algorithm, we have 
$F(c) = \frac 12\| A^{-1}\diag(c) u - g\|_2^2$, and $G(c) = \lambda\|c\|_1$. As shown in \cite{ipiano}, 
the gradient of $F$ with respect to $c$ is given as:
\[
\nabla F(c) = \diag(-(\cI + L)u + g)(A^\top)^{-1} (u - g)\,.
\]

For smooth TV regularization, $F(c) = \frac 12\| u(c) - g\|_2^2$, $u(c)$ is the solution of the 
lower-level TV regularized inpainting model. In order to calculate the gradient of $F$ with respect to $c$, we can make use 
of the implicit differentiation technique, see \cite{ChenGCPR13} for more details. The gradient is given as
\[
\nabla F(c)\big|_{c^*} = - D_c(u^*,c^*) (D_u(u^*,c^*))^{-1}(u^* - g) \,,
\]
where $u^*$ is the optimal solution of the lower-level problem in \eqref{general} at point $c^*$. As stated in 
\cite{ChenGCPR13}, in order to get an accurate gradient $\nabla F(c)$, we need to solve the lower-level problem as 
accurately as possible. To that end, we exploit Newton's method to solve the lower-level problem. 

Now we can make use of iPiano to solve this optimization problem. The algorithm is summarized below:  

\begin{algorithm}\caption{iPiano for solving problem \eqref{saddlepoint}}\label{algo2}
\begin{itemize}
\item [(1)] 
Choose $\beta \in [0,1)$, $l_{-1}>0$, $\eta > 1$, and 
initialize $c^0 = 1$ and set $c^{-1}= c^0$.
\item[(2)]  Then for $n\ge 0$, conduct a line search to find the smallest nonnegative integers $i$ such that 
with $l_n = \eta^i l_{n-1}$, the following inequality is satisfied 
\begin{align}\label{check}
      F(c^{n+1}) \leq F(c^n) + \scal{\nabla F(c^n)}{c^{n+1}-c^n} \nonumber\\
      + \frac{l_{n}}{2}\norm[2]{c^{n+1} - c^n}^2 \,,
    \end{align}
where $c^{n+1}$ is calculated from \eqref{rule} by setting $\beta = 0$. \\
Set $l_n = \eta^i l_{n-1}$, $\alpha_n < 2(1-\beta)/l_n$, and compute 
\begin{equation}\label{rule}
      c^{n+1} = (I+\alpha_n \partial G)^{-1}(c^n-\alpha_n \nabla F(c^n) +
      \beta(c^n-c^{n-1}))\,. 
\end{equation}
\end{itemize}
\end{algorithm}

\section{Numerical experiments}
In this section, we first discuss how to choose an efficient algorithm 
for solving the model \eqref{general} for different cases. Then we investigate the inpainting performance 
for different models under the unified assumption that we only make use of 5\% pixels. 
All the experiments were conducted on a server with Intel X5675 processors (3.07GHz), and all the investigated 
algorithms were implemented in pure Matlab code. We exploited three different test images 
(``Trui'', ``Walter'' and ``Peppers'') which are also 
used in previous works \cite{MainbergerSpatialTonal, HoeltgenSW13}.
\begin{figure*}[t!]
  \begin{center}
    \subfigure[Trui]{\includegraphics[width=0.22\textwidth]
{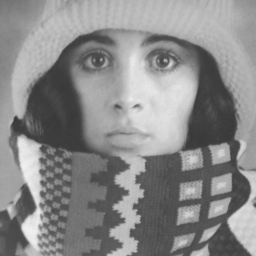}}
    \subfigure[Peppers]{\includegraphics[width=0.22\textwidth]
{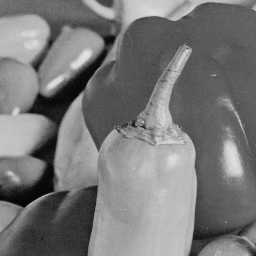}}
    \subfigure[Walter]{\includegraphics[width=0.22\textwidth]
{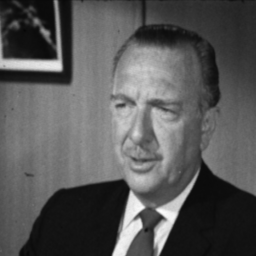}}
    \subfigure[Lena]{\includegraphics[width=0.22\textwidth]
{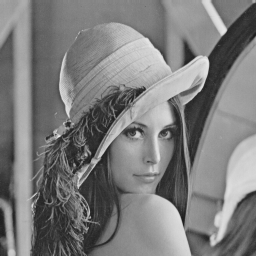}}
    \caption{Four test images used in our experiments}
\label{testimages}
  \end{center}
\end{figure*}

\subsection{Implementation details}
In our implementation, the parameter $\gamma$ of preconditioning technique is chosen as 
$\gamma = 10^{-6}$. For the SPPD algorithm, the parameter $\mu_1$ and $\mu_2$ are set as follows: 
(1)~for the Laplacian interpolation based compression model, $\mu_1 = 0.05,~\mu_2 = 0$; (2) for 
biharmonic interpolation based model, $\mu_1 = 0.1,~\mu_2 = 0.2$; and (3) for smoothed TV based 
model, ${\mu_1 = 0.05,~\mu_2 = 0.1}$. The set $\cC$ is defined in the range of $[0,~c_{max}]$ with 
$c_{max} = 1 - 10^{-6}$.

For the iPiano algorithm, we make use of the following parameter settings: 
\[
l_{-1}= 1,~\eta = 1.2,~\beta = 0.75,~\alpha_n = 1.99(1-\beta)/l_n \,.
\]
In order to exploit possible larger step size in practice, we use the
following heuristic: 
If the line search inequality \eqref{check} is fulfilled, we decrease the evaluated Lipschitz constant $L_n$ slightly 
by using a factor $1.02$, i.e., setting $l_n = l_n / 1.02$.

\subsection{Choosing appropriate algorithm for each individual model}
For Laplacian interpolation based compression model, we found that 
when using the proposed preconditioning technique, the required iterations can be reduced to about 150 outer iterations 
and 2000 inner iterations, which is a tremendous decrease compared to prior work \cite{HoeltgenSW13}. 
However, for this problem, the iPiano algorithm can do better. Our
experiments show that usually 700 iterations are already enough to reach a lower energy. 
Concerning the run time, the SPPD algorithm needs about 2400s, but iPiano only takes about 622s. 
We conclude that iPiano is clearly a better choice for solving the Laplacian interpolation based compression model. 

Let us turn to the biharmonic interpolation based compression
model. Even though the linear operator is only slightly changed, when
compared to the Laplacian model, it turns out that the corresponding optimization problem becomes much harder to solve. 
The SPPD algorithm still works for this problem; however, as mentioned before, 
we have to introduce an additional penalty term on variable $u$,
otherwise the convergence behavior is very bad. 
Besides, we have to run the algorithm much longer, usually about 300 outer iterations and 4000 inner iterations. 
For the iPiano algorithm applied to this case, we have to significantly increase the amount of required iterations, typically, we 
have to run about 3500 iterations to reach convergence. 

For the biharmonic interpolation based compression model \eqref{newmask}, 
the difference between the results obtained by above two algorithms becomes more obvious. 
For instance, for the test image ``Trui'' with parameter $\lambda = 0.0028$, by using the SPPD algorithm, we arrive a 
final energy of 15.34; however, the final energy of iPiano is much lower, about 13.48, which basically implies that 
iPiano solves the corresponding optimization problem better. Concerning the run time, for this case, iPiano takes 
more computation time than Laplacian interpolation case. 
There are two reasons: (1) the amount of required iterations is increased by a factor of 5; (2) 
for iPiano, we have to solve two linear equation $A x = b$ and $A^\top x = b$ in each iteration and line search\footnote
{In our implementation we use the Matlab ``backslash'' operator.}, 
which becomes much more time consuming from Laplacian to biharmonic interpolation. 
Therefore, for this case, both algorithms show a similar runtime (about 5000s). 
Since iPiano achieves a lower energy with similar computational effort
this algorithm is preferable for the biharmonic model.

For the case of smoothed TV regularization, it becomes even harder to solve the lower-level 
problem and thus more time consuming. It is therefore advisable not to make use of iPiano. 
The SPPD algorithm is a better choice for this model. Solving 
smoothed TV regularization based model also needs about 5000s.

\subsection{Reconstruct an image only using $\sim$5\% pixels}
\begin{figure*}[t!]
  \begin{center}
    \subfigure[10\% random chosen data]{\includegraphics[width=0.19\textwidth]
{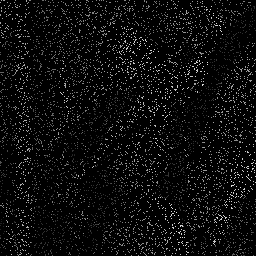}}
    \subfigure[Smoothed TV (276.37)]{\includegraphics[width=0.19\textwidth]
{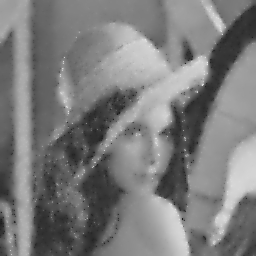}}
    \subfigure[Laplacian (244.48)]{\includegraphics[width=0.19\textwidth]
{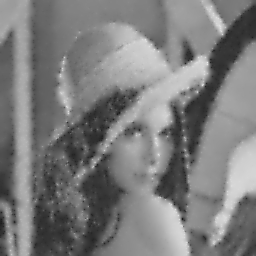}}
    \subfigure[Biharmonic (208.92)]{\includegraphics[width=0.19\textwidth]
{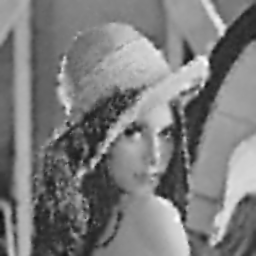}}
    \subfigure[Learned prior (165.01)]{\includegraphics[width=0.19\textwidth]
{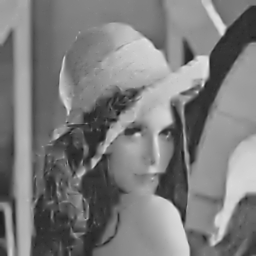}}
    \caption{Inpainting results of the degraded ``Lena'' image with 10\% randomly chosen pixels by using 
different methods. The number in the bracket is the resulting MSE. For randomly selected points, the inpainting model with learned MRF prior gives the best reconstruction result.}
\label{random}
  \end{center}
\end{figure*}
We evaluate the performance of three considered compression models based on three test images. 
For each individual model, we search optimal data points used for inpainting with the same amount of about 5\%, 
and then reconstruct an image by using these optimal points. In order to control the sparsity of selected data points to be 
5\% approximately, we have to carefully choose the parameter $\lambda$ for each model and for each processing image. 
The found optimal mask $c$ is continuous, and then we binarize it by a threshold parameter $\eps_T = 0.01$. 

Concerning the measurement of reconstruction quality, 
we make use of the mean squared error (MSE) to keep consistent with previous work, which is given by 
\[
MSE(u, g) = \frac 1 N \sum_{i=1}^{N} (u_i - g_i)^2 \,.
\]
The MSE is computed with the assumption that the image gray value is in the range of [0,~255]. 
As shown in previous work \cite{HoeltgenSW13}, for Laplacian interpolation, 
it is straightforward to consider an additional post-processing step, which is 
called gray value optimization (GVO) to further improve the reconstruction quality. 
We also consider this strategy for Laplacian and biharmonic interpolation after binarising the mask $c$, 
which is formulated as following optimization problem
\begin{equation}\label{gvo}
\arg \min_{x \in \R^M} \|A^{-1}S^\top x - g\|^2_2 \,,
\end{equation}
where $A$ is defined in the same way as in \eqref{newmask}. 
$S \in \R^{M \times N}$ is the sampling matrix derived from the diagonal matrix $\diag(c)$ by deleting the rows whose 
elements are all zero. $M$ is the number of points in the mask $c$ with a value of 1. Obviously, \eqref{gvo} is a least squared 
problem, which has the closed form solution
\[
x = \left(S (A^\top)^{-1}A^{-1}S^\top\right)^{-1}S(A^\top)^{-1}g.
\]
However, in practice it turns out that this computation is very time consuming because we have to calculate $A^{-1}$ explicitly. 
Therefore, we turn to L-BFGS algorithm to solve this quadratic optimization problem. 

For smoothed TV regularization model, we also consider this GVO post-processing step, which is given by the 
following bi-level optimization problem
\begin{align}\label{gvoTV}
  \min_{x \in \R^M} & l(x) = \frac{1}{2}\|u(x) - g\|_2^2\\
  \text{s.t.}  \;  & u(x) = \arg \min_{u} \|\nabla u\|_\eps + \frac 1 2 \|B^{\frac 1 2} (u - S^\top x)\|^2_2\,, 
\nonumber
\end{align}
where the sampling matrix $S \in \R^{M \times N}$ is the same as in \eqref{gvo}. We also make use of L-BFGS to solve this 
problem. To that end, we need to calculate the gradient of $l$ with respect to $x$, which is given as
\[
\nabla l(x)\big|_{x^*} = - D_x(u^*,x^*) (D_u(u^*,x^*))^{-1}(u^* - g) \,,
\]
where $D_u$ is the Hessian matrix given in \eqref{hessian}, $D_x = -SB$, $u^*$ is the solution of 
the lower-level problem at point $x^*$.

We summarize the results in Figure~\ref{comparisonfigure}. One can see that starting from the initial Laplacian interpolation 
based image compression model, we can achieve significant improvements of inpainting performance for all test images by using 
biharmonic interpolation based model, at the expense of computation time; 
however, switching to the smoothed TV regularization based model can not bring any 
improvement even with more computation time. 
To the best of our knowledge, concerning the inpainting performance of the biharmonic interpolation model, 
it is the first time to achieve such an accurate reconstruction by using only 5\% pixels. 

\section{Conclusion and future work}
In this paper, we extended the Laplacian interpolation based image compression model to 
more general inpainting based compression model. 
Starting from the Laplacian interpolation, 
we investigated two variants, namely biharmonic interpolation and smoothed TV regularization inpainting model, 
to improve the compression performance. 
In order to solve the corresponding optimization problems efficiently, we introduced two fast algorithms: 
(1) successive preconditioning primal dual algorithm and (2) a recently proposed non-convex optimization algorithm - 
iPiano. Based on these algorithms, for each model, 
we found the most useful 5\% pixels, and then reconstructed an image from the optimal data. 
Numerical results demonstrate that (1) biharmonic interpolation gives the best reconstruction performance and 
(2) the smoothed TV regularization model can not generate superior
results over the Laplacian interpolation method. 

Future work consists of two aspects: (1) more efficient algorithm to solve the corresponding optimization problems. Even though 
the introduced algorithms are fast, they are still very time consuming for complicated models, e.g., 
biharmonic interpolation and smoothed TV regularization models. (2) exploiting more sophisticated inpainting models to further improve 
the compression model. A possible candidate is to make use of the
inpainting model with a learned MRF prior
\cite{ChenGCPR13, chenTIP2013}, which is shown to work well for image inpainting with randomly selected points. Figure~\ref{random} 
presents an example to show the inpainting performance of the learned model for randomly selected data points. 
One can see that in this random case, the inpainting model with learned MRF prior can generate the best result, 
and therefore, we believe that it can achieve better result for image compression.

\begin{figure*}[t!]
  \begin{center}
    \subfigure[4.98\%]{\includegraphics[width=0.2\textwidth]
{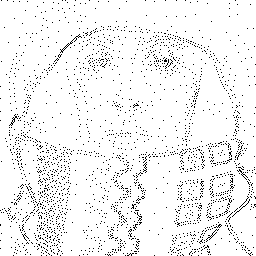}}
    \subfigure[4.98\%]{\includegraphics[width=0.2\textwidth]
{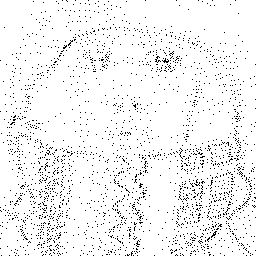}}
    \subfigure[6.90\%]{\includegraphics[width=0.2\textwidth]
{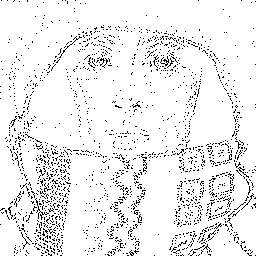}}
    \subfigure[4.95\%]{\includegraphics[width=0.2\textwidth]
{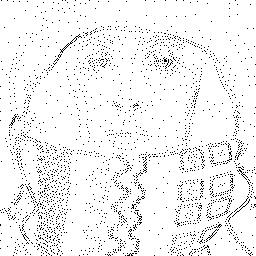}}\\ \vspace{-0.2cm}
    \subfigure[MSE: 16.89]{\includegraphics[width=0.2\textwidth]
{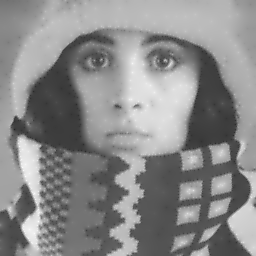}}
    \subfigure[MSE: 10.60]{\includegraphics[width=0.2\textwidth]
{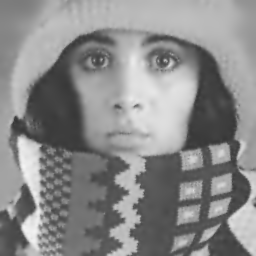}}
    \subfigure[MSE: 17.98]{\includegraphics[width=0.2\textwidth]
{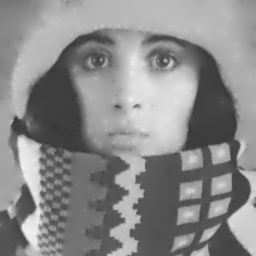}}
    \subfigure[MSE: 16.95]{\includegraphics[width=0.2\textwidth]
{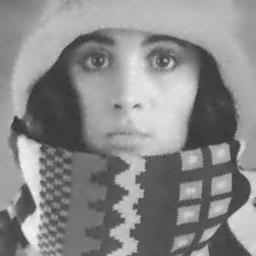}}\\ \vspace{-0.2cm}
    \subfigure[4.84\%]{\includegraphics[width=0.2\textwidth]
{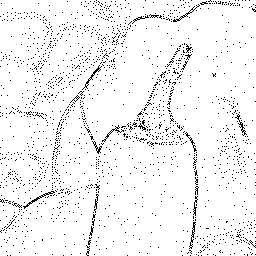}}
    \subfigure[4.89\%]{\includegraphics[width=0.2\textwidth]
{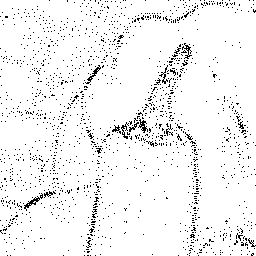}}
    \subfigure[5.69\%]{\includegraphics[width=0.2\textwidth]
{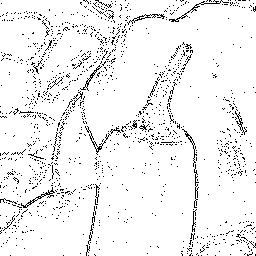}}
    \subfigure[5.02\%]{\includegraphics[width=0.2\textwidth]
{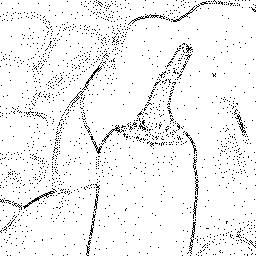}}\\ \vspace{-0.2cm}
\subfigure[MSE: 18.99]{\includegraphics[width=0.2\textwidth]
{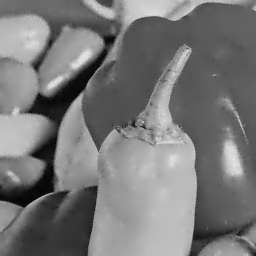}}
 \subfigure[MSE: 17.81]{\includegraphics[width=0.2\textwidth]
{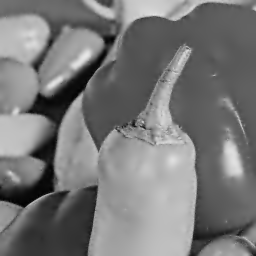}}
\subfigure[MSE: 21.14]{\includegraphics[width=0.2\textwidth]
{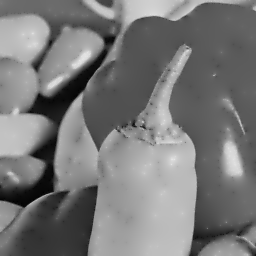}}
\subfigure[MSE: 18.44]{\includegraphics[width=0.2\textwidth]
{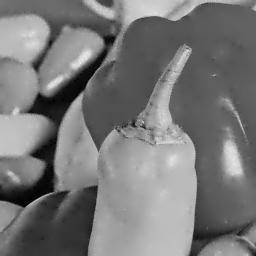}}\\ \vspace{-0.2cm}
    \subfigure[4.82\%]{\includegraphics[width=0.2\textwidth]
{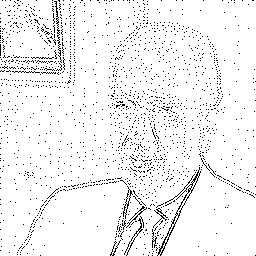}}    
\subfigure[4.59\%]{\includegraphics[width=0.2\textwidth]
{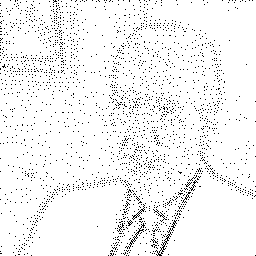}}
\subfigure[5.86\%]{\includegraphics[width=0.2\textwidth]
{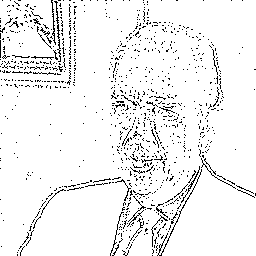}}
\subfigure[5.00\%]{\includegraphics[width=0.2\textwidth]
{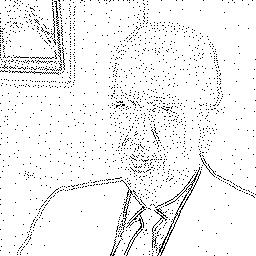}}\\ \vspace{-0.2cm}
    \subfigure[MSE: 8.03]{\includegraphics[width=0.2\textwidth]
{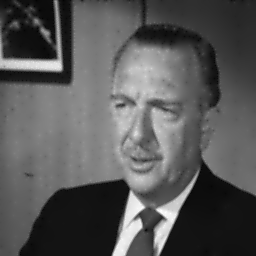}}   
    \subfigure[MSE: 4.85]{\includegraphics[width=0.2\textwidth]
{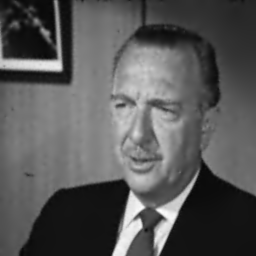}}
    \subfigure[MSE: 10.52]{\includegraphics[width=0.2\textwidth]
{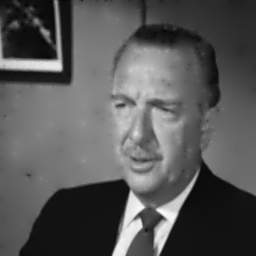}}
\subfigure[MSE: 7.59]{\includegraphics[width=0.2\textwidth]
{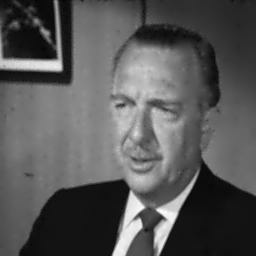}}
    \caption{Image inpainting results by using approximate 5\% pixels. The interpolation data points used for reconstruction is masked in 
black. The continuous mask $c$ is binarized with a threshold parameter $\eps_T = 0.01$. 
\textbf{From left to right}: 
\textbf{(1)} optimal mask found with Laplacian interpolation and the corresponding recovery image by using the optimal data points, 
\textbf{(2)} results of biharmonic interpolation model, 
\textbf{(3)} results of smoothed TV regularization approach, 
\textbf{(4)} results of \cite{HoeltgenSW13}}
\label{comparisonfigure}
  \end{center}
\end{figure*}
\bibliographystyle{cmpproc}
\bibliography{compression}

\end{proceeding}
\end{document}